\documentclass[journal,transmag]{elsarticle}

\usepackage{setspace}
\usepackage{amssymb,amsmath,amsthm}
\usepackage[margin=0.9in]{geometry} 
\usepackage{verbatim} 
\usepackage{listings} 
\usepackage{color}
\usepackage[normalem]{ulem}

\newcommand{\bb}{\mathbf}

\usepackage{tikz}
\usetikzlibrary{arrows}
\usepackage{pgfplots}

\usepackage[]{algorithm2e}

\newproof{pf}{Proof}

\begin{document}
	
\begin{frontmatter}

	\title{On the Computation and Applications of Large Dense Partial Correlation Networks
	}


\author[newhavenaddress]{Keith Dillon\corref{mycorrespondingauthor}}
\cortext[mycorrespondingauthor]{Corresponding author. Tel. 1-949-478-1736}
\ead{kdillon@newhaven.edu}

\address[newhavenaddress]{Department of Electrical and Computer Engineering and Computer Science, \\ University of New Haven, West Haven, CT, USA}


\begin{abstract}

While sparse inverse covariance matrices are very popular for modeling network connectivity, the value of the dense solution is often overlooked. 
In fact the $\ell_2$-regularized solution has deep connections to a number of important applications to spectral graph theory, dimensionality reduction, and uncertainty quantification. 
We derive an approach to directly compute the partial correlations based on concepts from inverse problem theory.
This approach also leads to new insights on open problems such as model selection and data preprocessing, as well as new approaches which relate the above application areas. 

\end{abstract}

\end{frontmatter}



\section{Introduction}

An important application of networks, beyond merely describing relationships between variables, is to make quantitative predictions of variable behavior. 
Gaussian graphical models \cite{whittaker_graphical_2009} are a popular approach to describing networks, and are directly related to variable prediction via linear regression \cite{pourahmadi_covariance_2011}.
The focus is often on graphical model edges described by partial correlations which are zero, identifying pairs of nodes which are conditionally independent \cite{baba_partial_2004}. 
For example, the graphical LASSO \cite{friedman_sparse_2008} imposes a sparse regularization penalty on the precision matrix estimate, seeking a network which trades off predictive accuracy for sparsity.
This provides a network which more interpretable and efficient to use, however it is not clear that sparse solutions actually generalize better to new data than dense solutions do \cite{williams_back_2018}.

Meanwhile, a different research direction is based on forming edges via some simple relationship such as affinity or univariate correlation. This limited network is used as a starting point for computing sophisticated dense estimates of relatedness between nodes, providing a deeper analysis of network structure.
In such research, sparsity is usually imposed on the simple network, however the subsequent analysis is often based on methods which inherently presume Gaussian statistics and $\ell_2$ penalties in some sense. 
A popular method in this category is spectral clustering \cite{von_luxburg_tutorial_2007}, which has been used to solve problems in a range of areas such as image processing \cite{weiss_segmentation_1999}, graph theory \cite{saerens_principal_2004}, clustering on nonlinear manifolds \cite{belkin_laplacian_2003}, and brain parcellation \cite{dillon_spectral_2018}.
Spectral clustering is commonly described as a continuous relaxation of the normalized cut algorithm \cite{von_luxburg_tutorial_2007} for partitioning graphs.
In addition to approximately-optimal partitioning of graphs, other interpretations have been noted for spectral clustering such as random walks and finding smooth embeddings \cite{meila_spectral_2015}, sometimes involving minor variants of the algorithm, such as by normalizing the Laplacian differently \cite{meila_learning_2002}. 

In this paper we consider the estimation of partial correlations from the regularized regression direction. We show how the partial correlations can be computed as a scaled form of a quantity from inverse problems theory called the
%
%
%
%
%
%
%
%
resolution matrix \cite{jackson_interpretation_1972}, which quantifies the  information lost in an imaging system or process.
This matrix is often viewed as an approximate identity matrix or blurring kernel \cite{berryman_analysis_2000}. 
By inspecting the resolution matrix which describes an imaging system, we can quantify the resolution attainable when reconstructing an image from data collected by the system.
This has been used for a number of years in fields such as geophysical \cite{ganse_uncertainty_2013} and optical \cite{boas_improving_2004} imaging. 
Researchers in these fields have developed a variety of efficient numerical computation and approximation methods for the resolution matrix  \cite{nolet_explicit_1999,maccarthy_efficient_2011,trampert_resolution_2013,zhang_estimation_1995,minkoff_computationally_1996,soldati_global_2006}.
We will show how Gaussian graphical models, as well as spectral methods, can be viewed as variations on the resolution matrix
which quantifies how well we can differentiate each variable from the rest with the given dataset. 
We start by reviewing the network estimation problem via Gaussian graphical models. 
Then we analyze this problem via the resolution matrix and show its relation to the Gaussian graphical model.  
Next we show how the resolution matrix can be related to spectral methods and as a result we can produce new methods combining partial correlation and spectral embeddings. 

\section{Theory}
%
%
%
%
%
%
%
%
%
%
%
%

\begin{figure}[h!] \centering 
	\scalebox{0.45}{\includegraphics[clip=true, trim=0in 0in 0in 0in]{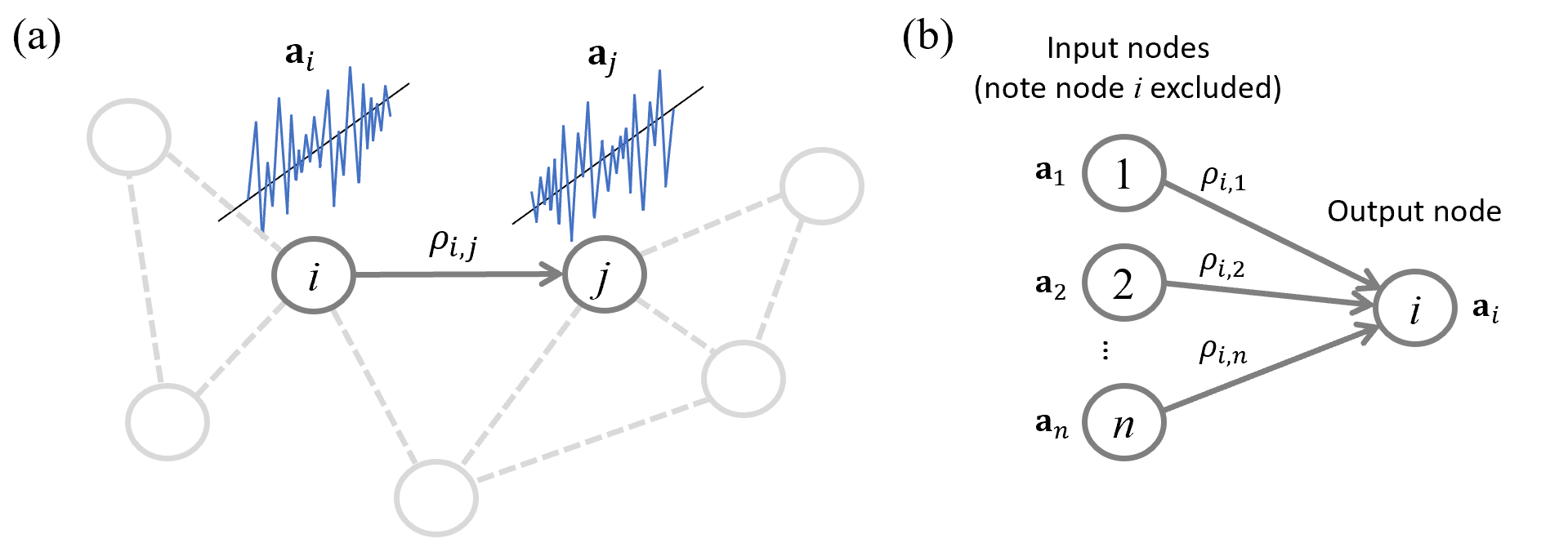}} 
	\caption{(a) Depiction of network, where the partial correlation $\rho_{i,j}$ provides the edge weight between nodes $i$ and $j$; the vectors $\bb a_i$ and $\bb a_j$ contain samples of the random variables $a_i$ and $a_j$, respectively, describing samples of the variables corresponding to nodes. (b) The network neighborhood refers to the subnetwork consisting of inputs to the $i$th node.}
	\label{network0}
\end{figure}

We start by reviewing the use of partial correlation to estimate connectivity (as depicted in Fig. \ref{network0}), and its estimation via regression coefficients.
We model the signal at the $i$th node as the zero-mean Gaussian random variable $a_i$.
The partial correlation $\rho_{i,j}$ between $a_i$ and $a_j$, is the Pearson correlation between the residuals after regressing out all other nodes except $i$ and $j$ from each.
Rather than directly performing this computationally-intensive calculation on data, there are two general categories of methods used.
The first category exploits the relationship with the precision matrix, the inverse of the sample covariance matrix. 
The second category, which we will consider here, exploits the relationship between the partial correlation and the regression coefficients for the neighborhood selection problem \cite{meinshausen_high-dimensional_2006}.
These regression coefficients are defined as the solutions $\beta_{i,j}$ to the linear system
\begin{align}
a_i = \sum_{k \neq i} \beta_{i,k} a_k + \epsilon_i,
\label{regression0}
\end{align}
where $a_i$ is the $i$th variable and $\epsilon_i$ is the residual.
From these $\beta_{i,j}$ we can estimate $\rho_{i,j}$ as  \cite{pourahmadi_covariance_2011},
\begin{align}
\rho_{i,j} &= -\beta_{i,j} \sqrt{\frac{\sigma^{i,i}}{\sigma^{j,j}}}, 
\label{rv_rho}
\end{align}
using the residual variances $\sigma^{i,i} = (Var(\epsilon_i))^{-1}$.
A common alternative formulation exploits the symmetry of the partial correlation (i.e., that $\rho_{i,j} = \rho_{i,j}$ by definition) and use the geometric mean to cancel the residual variances as in \cite{schafer_shrinkage_2005}
\begin{align}
	\rho_{i,j} &= \text{sign}(\beta_{i,j})  \sqrt{\beta_{i,j}\beta_{j,i}}.
	\label{rv_rho_geomean}
\end{align}
This also has the advantage of enforcing symmetry in sample estimates.
If the signs of  $\beta_{i,j}$ and $\beta_{i,j}$ differ, $\rho_{i,j}$ is typically set to zero.

To write matrix equations for the sample estimates, we define $\bb A$ as a matrix containing data, with samples for  $a_i$ in the $i$th column $\bb a_i$ (which we will assume has been standardized).
The regression problem of Eq. (\ref{regression0}) becomes the linear system
\begin{align}
\bb a_i = \bb A_{-i} \boldsymbol{\beta}_i + \boldsymbol{\epsilon}_i,
\label{regression1}
\end{align}
where $\bb A_{-i}$ is the matrix $\bb A$ with the $i$th column set to zeros; the vector $\boldsymbol{\beta}_i$ is the estimates of the regression coefficients, where  $(\boldsymbol{\beta}_i)_j$ is the estimate of $\beta_{i,j}$  (setting $\beta_{i,i}=0$); and $\boldsymbol{\epsilon}_i$ is the vector of samples of the residual $\epsilon_i$.
The least-squares solution for the regression coefficients is the minimizer
\begin{align}
\boldsymbol{\beta}_i = \arg \underset{\boldsymbol{\beta}'_i}{\min}\Vert \bb A_{-i} \boldsymbol{\beta}'_i - \bb a_i \Vert_2^2.
\label{eq_neighborhood_reg}
\end{align}
This can be computed via the pseudoinverse, $\bb A^\dagger_{-i} = \bb A_{-i}^T(\bb A_{-i}^T \bb A_{-i})^{-1}$, giving 
\begin{align}
\boldsymbol{\beta}_i = \bb A_{-i}^\dagger \bb a_i.
\label{pseudo}
\end{align} 
%
%
%
The sample residual variances are then
\begin{align}
d_i^2 &= 
\Vert \boldsymbol{\epsilon}_i \Vert^2 = \Vert\bb A_{-k}^\dagger \boldsymbol{\beta}_i -  \bb a_i \Vert^2
\end{align} 
which we form into a vector $\bb d$ with $(\bb d)_i = d_i$.
With this we can write the sample version of Eq. (\ref{rv_rho}) as 
\begin{align}
\bb P = \bb D_{\bb d} \bb B \bb D_{\bb d}^{-1},
\label{P_DBD}
\end{align}
where $\bb D_{\bb d}$ is a diagonal matrix with $D_{i,i} = d_i$, and we have formed $\bb B$ with $\boldsymbol{\beta}_i$ as columns.
$\bb P$ contains our sample-based estimates of the partial correlations, with $P_{i,j}$ describing the partial correlation between nodes $i$ and $j$.
Again, we can avoid calculating the residual variances as in Eq. (\ref{rv_rho_geomean}) as follows,
\begin{align}
\bb P =  \text{sign}(\bb B) \odot (\bb B \odot \bb B^T)^{\circ \frac{1}{2}},
\label{P_BBB}
\end{align}
using the Hadamard product $\odot$ and element-wise exponential $\circ\frac{1}{2}$, and where the sign function is taken element-wise.


The above formulation is general in that a pseudoinverse exists for both overdetermined matrices (more rows than columns, implying more samples than variables) as well as underdetermined matrices. And indeed when the matrix is invertible, the pseudoinverse reduces to the matrix inverse. 
In this paper, our focus will be on cases where the regression problem of Eq. (\ref{eq_neighborhood_reg}) is ill-posed and the matrix $\bb A$ is underdetermined or rank-deficient.  
In this scenario, there are infinite possible solutions for the regression coefficients.
A popular approach to address this is to directly replace the regression coefficients with regularized versions, such as via Ridge Regression \cite{kramer_regularized_2009}. 
In such a case, we'd incorporate a penalty term into Eq. (\ref{eq_neighborhood_reg})
\begin{align}
\boldsymbol{\beta}_i = \arg \underset{\boldsymbol{\beta}'_i}{\min}\Vert \bb A_{-1} \boldsymbol{\beta}'_i - \bb a_i \Vert_2^2 + \lambda \Vert \boldsymbol{\beta}'_i \Vert_2^2,
\label{eq_neighborhood_reg_rr}
\end{align}
introducing a regularization parameter $\lambda$, chosen via cross-validation, for example.
We describe the solution to this problem using a regularized version of the pseudoinverse, 
\begin{align}
\boldsymbol\beta_i &= (\bb A_{-i})_\lambda^\dagger \bb a_i \\
(\bb A_{-i})^\dagger_\lambda &= \bb A_{-i}^T(\bb A_{-i}^T \bb A_{-i} + \lambda \bb I)^{-1}.
\end{align}

\subsection{The Resolution Matrix}

The regression problem of Eq. (\ref{eq_neighborhood_reg_rr}) is unique to each node $i$, which implies a large computational effort to solve for the entire network.
However a closely-related problem which is much easier to solve can be formed by  allowing self-loops in the network. 
The regression system for the network with self-loops is simply
\begin{align}
\bb a_i = \bb A \boldsymbol{\beta}_i^{(0)} + \boldsymbol{\epsilon_i}
\label{regression_loops0}
\end{align}
which differs from Eq. (\ref{regression1}) in that $\bb A_{-i}$ is replaced with $\bb A$, the original data matrix without excluding the $i$th column.
The solution to Eq. (\ref{regression_loops0}) is 
\begin{align}
\boldsymbol{\beta}_i^{(0)} = \bb A^{\dagger} \bb a_i.
\end{align}
If we form the matrix $\bb B^{(0)}$ with $\boldsymbol{\beta}_i^{(0)}$ as its $i$th column, we get 
\begin{align}
\bb B^{(0)} = \bb A^{\dagger} \bb A = \bb R
\end{align}
Where we have defined the resolution matrix $\bb R$ for the system \cite{dillon_spectral_2018}.
In an inverse problem, this matrix can be viewed as a blurring matrix or approximate identity, which describes the loss of information between the unknown input, and the measured output. 
In our case, the ``input'' is the unknown regression coefficients, and the output is the signal $\bb a_i$. 
The characteristics of the dataset $\bb A$, described by the resolution matrix, determine how well we can use the data $\bb a_i$ to estimate the true $\boldsymbol\beta_i$.
%
%
%
%
%
%
%
If we compute the sample covariance matrix as (assuming standardized columns for $\bb A$)
\begin{align}
\bb C = \frac{1}{m}\bb A^T \bb A,
\end{align}
Then we can equivalently compute the network resolution matrix as
\begin{align}
\bb R = \bb C^\dagger \bb C.
\end{align}
The covariance, or some sparse approximate version, is often used to define a weighted adjacency matrix.
Hence $\bb R$ can equivalently be calculated directly from the adjacency matrix of a network, rather than from the data used to estimate connectivity.  
In the random walk perspective, we can view the resolution matrix as the resolution of our knowledge regarding the next step in the walk. 
%
%

%
%
%

\subsubsection{Regularization of $\bb R$}

Earlier we noted the use of regularization techniques in partial correlation estimates by utilizing a regularized version of the regression estimate. 
In image science there is a parallel concept involving the regularization of the resolution matrix, where the goal is a more robust estimate of the system resolution. 
The most useful approach to regularizing the  resolution matrix is the product of a matrix with a regularized version of its generalized inverse \cite{an_simple_2012}. 
In our case this would be 
\begin{align}
\bb R_{\lambda} = \bb A_\lambda^\dagger \bb A
\end{align}
where $\bb A_\lambda^\dagger$ is the regularized generalized inverse, which we can write as
\begin{align}
\bb A_\lambda^\dagger &=  (\bb A^T \bb A + \lambda \bb I)^{-1}\bb A^T,
\label{eq_pinv_OD}
\end{align}
with regularization parameter $\lambda$.
It can be useful to also use the form 
\begin{align}
\bb A_\lambda^\dagger &=  \bb A^T(\bb A \bb A^T + \lambda \bb I)^{-1}.
\label{eq_pinv_UD}
\end{align}
It can be seen that the form in Eq. (\ref{eq_pinv_OD}) is equal to Eq. (\ref{eq_pinv_UD}) by using the singular value decomposition (SVD) $\bb A = \bb U \bb S \bb V^T$, and showing that both versions of the regularized pseudoinverse can be written as 
\begin{align}
\bb A_\lambda^\dagger &=\bb V \bb S^T_\lambda \bb U^T,
\end{align}
where $\bb S$ is the $m \times n$ diagonal matrix of singular values $S_{i,i} = \sigma_i$, and $\bb S_\lambda$ is a $m \times n$ diagonal matrix with attenuated singular values, $(\bb S_\lambda)_{i,i} = \sigma_i (\sigma_i^2+\lambda)^{-1}$.
Eq. (\ref{eq_pinv_OD}) requires a $n \times n$ matrix inversion while Eq. (\ref{eq_pinv_UD}) requires a $m \times m$ matrix inversion.


An alternative form of regularization used in many applications is truncation of the singular value decomposition of the data. 
This is more often viewed as a form of dimensionality reduction and a preprocessing step, but is closely-related to $\ell_2$-regularization, as can be seen by writing a regularized pseudoinverse of the form
\begin{align}
\bb A_r^\dagger &=\sum_{i=1}^{r} \frac{1}{\sigma_i}\bb v_i \bb u_i^T \\
&=\bb V \bb S^T_t \bb U^T,
\label{eq_SVD_trunc_pinv}
\end{align}  
with $(\bb S_t)_{i,i} = \sigma_i^{-1}$ for $i\le r$ and zero otherwise.
So while $\ell_2$-regularization imposes shrinkage on the smaller singular values, SVD-truncation simply sets them to zero below some chosen threshold. 

\subsection{Using $\bb R$ to compute $\bb B$ and $\bb P$}

We will now relate the least-squares solution for $\bb A_{-i} \boldsymbol\beta_i = \bb a_i$, to the solution for $(\bb A_{-i}, \bb a_i)\, \boldsymbol\beta_i^{(0)} = \bb a_i$. For $\ell_2$-regularization, the solution for $\boldsymbol\beta_i$ is (using similar logic as that relating Eqs. (\ref{eq_pinv_OD}) and (\ref{eq_pinv_UD}))
\begin{align}
\boldsymbol\beta_i 
&= \bb A_{-i}^T \left( \bb A_{-i} \bb A_{-i}^T + \lambda \bb I\right)^{-1} \bb a_i \\
&= \bb A_{-i}^T \left( \bb A \bb A^T + \lambda \bb I - \bb a_i \bb a_i^T\right)^{-1} \bb a_i
\label{eq_pinv_both}
\end{align}
We employ the matrix inversion lemma, 
\begin{align}
\left( \bb M + \bb U \bb C \bb V\right)^{-1} = \bb M^{-1} - \bb M^{-1} \bb U \left( \bb C^{-1} + \bb V \bb M^{-1} \bb U\right)^{-1} \bb V \bb M^{-1},
\end{align}
with $\bb M = \bb A \bb A^T + \lambda \bb I$, $\bb U = \bb a_i$, $\bb C = -1$ and $\bb V = \bb a_i^T$.
Plugging these in gives
\begin{align}
\boldsymbol\beta_i 
&= \bb A_{-i}^T(\bb A \bb A^T + \lambda \bb I)^{-1}\bb a_i - 
\frac{\bb A_{-i}^T(\bb A \bb A^T + \lambda \bb I)^{-1} \bb a_i\bb a_i^T (\bb A \bb A^T + \lambda \bb I)^{-1}\bb a_i}{ -1 + \bb a_i^T (\bb A \bb A^T + \lambda \bb I)^{-1} \bb a_i} 
\label{eq_beta_i_lemma}
\end{align}
Meanwhile, the least-squares solution for $\boldsymbol\beta_i^{(0)}$ is, in the underdetermined case employing Ridge regression,
\begin{align}
\boldsymbol\beta_i^{(0)} 
&= \bb A^T \left( \bb A \bb A^T + \lambda \bb I\right)^{-1} \bb a_i \\
&=  
\begin{pmatrix}
\bb A_{-i}^T \\
\bb a_i^T
\end{pmatrix}
( \bb A \bb A^T + \lambda \bb I)^{-1}  \bb a_i \\
&= \begin{pmatrix}
\bb r_{-i}^{(\lambda)} \\
R_{i,i}^{(\lambda)}
\end{pmatrix},
\end{align}
where we have  defined $\bb r^{(\lambda)}_{-i}$ as the $i$th column of the regularized resolution matrix with the $i$th element removed, and $ R^{(\lambda)}_{i,i}$ as the $(i,i)$th element.
We have also assumed the $i$th column is the last column for convenience. 
Utilizing these definitions in Eq. (\ref{eq_beta_i_lemma}) gives,
\begin{align}
\boldsymbol\beta_i= \left(\frac{1}{1-R_{i,i}^{(\lambda)}} \right)
\bb r_{-i}^{(\lambda)}  
\end{align}
So the $i$th regression coefficients are a scaled version of the resolution matrix with the $i$th element removed. 
Further, the scalar is simply calculated from the $i$th element itself.

Regularization by SVD-truncation proceeds along similar logic, but must be viewed  differently. 
In this case it regularization is applied as a preprocessing step prior to network estimation, whereas $\ell_2$-regularization is assumed to be applied as a penalty during the regression calculation.
So for SVD-truncation, given a dataset $\bb A$, we form the SVD-truncated version,  
\begin{align}
  \bb A_r = \bb U \bb S_r \bb S_V^T,
  \label{eq_svd_trunc}
\end{align}
where $\bb S_r$ is the matrix of singular values, with the values beyond the $r$ set to zero (when sorted in decreasing order).
If we compute the pseudoinverse which yields the least-length solution and insert the truncated decomposition from Eq. (\ref{eq_svd_trunc}), we get
\begin{align}
\bb A_r^\dagger &=  \bb A_r^T(\bb A_r \bb A_r^T)^{-1} \\
&= \bb V \bb S_t^T \bb U^T,
\label{eq_SVD_pinv_UD}
\end{align}
which is the pseudoinverse form of Eq. (\ref{eq_SVD_trunc_pinv}).
Noting that Eq. (\ref{eq_SVD_pinv_UD}) is just the underdetermined version in Eq. (\ref{eq_pinv_both}) with $\lambda=0$, we can proceed along the same steps to relate the regression coefficients to the resolution matrix.

To write the matrix version of the relation between $\bb B$ and $\bb R$, 
we form the matrix $\bb R_{-\bb d}$ defined as $\bb R$ with the values on the diagonal set to zero, and perform the scaling,
\begin{align}
\bb B =\bb R_{-\bb d} \bb D_{\bb s}
\label{B_RD}
\end{align}
where $\bb D_{\bb s}$ is a diagonal matrix with $s_i = (1-R_{i,i})^{-1}$ on the diagonal.
Then, combining Eqs. (\ref{P_DBD}) and (\ref{B_RD}) we get
\begin{align}
\bb P = \bb D_{\bb d} \bb R_{-d} \bb D_s \bb D_{\bb d}^{-1}.
\label{P_DBD}
\end{align}
In cases where $\bb P$ and $\bb R$ are too large to fit in memory, we can compute columns on the fly as,
\begin{align}
\bb p_i &= \frac{1}{d_i (1-R_{i,i})}\bb D_d \bb r_{-i} \\
&= \frac{1}{d_i (1-R_{i,i})}\bb I_{-i}\bb D_d \bb A^\dagger \bb a_i,
\label{p_i_asymmetric}
\end{align}
where $\bb I_{-i}$ is the identity matrix with a  zero in the $(i,i)$th position.
This requires that we calculate and store a regularized pseudoinverse of our data matrix, which is of the same size as our original data matrix.
Additionally we can pre-compute the diagonal of $\bb R$ and the $\bb d$ vector.
The diagonal of $\bb R$ is equal to the sum of its eigenvectors squared, so can be computed as a by-product of truncated-SVD regularization.
In extremely-large-scale situations, the diagonal
can be computed using even more efficient techniques 
\cite{maccarthy_efficient_2011,trampert_resolution_2013}.
The $\bb d$ vector can then be computed via
\begin{align}
d_i &= \Vert \bb A \bb B - \bb A \Vert \\
&= \left\vert\tfrac{1}{1-R_{i,i}}\right\vert \left\Vert \bb A \left( \bb A^\dagger \bb a_{i} - \bb e_i\right)\right\Vert. 
\label{eq_di}
\end{align}

Alternatively, combining Eqs. (\ref{P_BBB}) and (\ref{B_RD}) we get the geometric mean formulation,
\begin{align}
\bb P 
&=  \text{sign}(\bb 1 \bb s^T) \odot (\bb s \bb s^T)^{\circ \frac{1}{2}} \odot \bb R_{-d}. 
\end{align}
To enforce the sign test, we set $P_{i,j}$ equal to zero when $\text{sign}(s_i) \ne \text{sign}(s_j)$ .
In cases where $\bb P$ and $\bb R$ are too large to fit in memory, we can compute columns on the fly as,
\begin{align}
\bb p_i &=  \text{sign}(s_i) \odot (s_i \bb s)^{\circ \frac{1}{2}} \odot \bb r_{-i} \\ 
&= \boldsymbol\sigma_i \odot (\bb A^\dagger \bb a_{i}).
\end{align}
where for convenience we have defined
\begin{align}
\boldsymbol\sigma_i = \begin{cases}
\text{sign}(s_i) \odot (s_i \bb s)^{\circ \frac{1}{2}}, &\text{if }  i\ne j, \text{ and } \text{sign}(s_i) = \text{sign}(s_j), \\
0, &\text{otherwise}.
\end{cases}
\end{align}

\section{Application}

Now we will consider the uses of the dense partial correlation estimates, where we demonstrate a close relationship to popular spectral methods.
%
%
%
In \cite{dillon_regularized_2017} we used distances between pairs of columns of $\bb R$ as a form of connectivity-based distance in a network clustering algorithm, and demonstrated how the distance could be computed efficiently even for large datasets where $\bb R$ could not fit in memory. 
The basic idea was to use the factors $\bb A^\dagger$ and $\bb A$, rather than computing $\bb R$ itself as follows,
\begin{align}
D^{(R)}_{i,j} &= \Vert \bb r_i - \bb r_j \Vert \\
&= \left\Vert \bb A^\dagger\left(  \bb a_i - \bb a_j \right) \right\Vert
\label{eq_R_dist}
\end{align}
In \cite{dillon_spectral_2018} we noted that if SVD-truncation was used as the regularization method, then the resolution matrix can be written as
\begin{align}
\bb R &= (\bb V \bb S_t^T \bb U^T) \bb U \bb S \bb V^T \\
      &= \bb V \bb I_r \bb V^T, 
\end{align}
where $\bb I_r$ is a truncated version of the identity, with zeros for the columns corresponding to discarded singular values.
Then the resolution distance can be written as 
\begin{align}
D^{(R)}_{i,j} &= \Vert \bb V \bb I_r \bb v^{(i)} -  \bb V \bb I_r \bb v^{(j)} \Vert \\
&= \left\Vert \bb I_r \left(  \bb v^{(i)} - \bb v^{(j)} \right) \right\Vert
\label{eq_R_dist}
\end{align}
where $\bb v^{(i)}$ is the $i$th row of $\bb V$ and so $\bb I_r \bb v^{(i)}$ is the $i$th row of the matrix formed by the first $r$ singular vectors of $\bb A$.
As these singular vectors of are the same as the singular vectors of the covariance matrix, this distance is a form of spectral embedding. 

We can similarly define conditional forms of spectral embedding and clustering using distances between columns of $\bb P$, which may be implemented as
\begin{align}
D_{i,j} &= \Vert \bb p_i - \bb p_j \Vert \\
&= \left\Vert \bb D_d\bb A^\dagger\left( \tfrac{1}{d_i (1-R_{i,i})} \bb a_i -   \tfrac{1}{d_j (1-R_{j,j})}\bb a_j \right) 
-\tfrac{R_{i,i}}{1-R_{i,i}} \bb e_i 
-\tfrac{R_{j,j}}{1-R_{j,j}} \bb e_j 
\right\Vert
\label{eq_P_dist}
\end{align}
where we have used the assymmetric version of $\bb P$ from Eq. (\ref{p_i_asymmetric}), and defined $\bb e_{i}$ as the $i$th column of the identity.
In dense networks with highly-collinear variables, we would expect the $R_{i,i}$ terms to be very small and hence the two trailing terms in Eq. (\ref{eq_P_dist}) to be negligible. 
Hence we find that the distance metric computed by taking the Euclidean distance between columns of $\bb P$ is essentially a weighted variation on a spectral distance metric.

We tested the use as a data embedding for supervised machine learning, by using the distance metric of Eq. (\ref{eq_P_dist}) in $k$-nearest neighbors classification.
We used a number of different standard test datasets from the UCI Machine Learning Repository \cite{dua_uci_2017}: the Iris dataset \cite{fisher_use_1936}, the Wine dataset, the Breast Cancer dataset, the Ionosphere dataset, and the Credit Approval dataset. 
We computed the accuracy at predicting class membership for each sample using others.
We used five-fold cross-validation, and computed accuracy as the average accuracy over the folds with the best choice of parameters for regularization and number of neighbors. 
Results comparing a standard distance metric (i.e., euclidean distance between feature vectors) and a partial correlation distance from Eq. (\ref{eq_P_dist}) are given in Table \ref{table_knn}.
We also tried optimizing a dimensionality reduction step for the base method, so that it would have an equal number of parameters to optimize over, but this did not improve the accuracy.
From Table 1, we find that the partial correlation distance performs either comparably or noticeably better, depending on the dataset.

%
%
%
%

\begin{table}
	\footnotesize
	\centering
	\caption{Classification with standard distance ($k$-NN) and partial correlation distance ($k$-PCN).}
	\label{table_knn}
	\begin{tabular}{  c  c   c  c  c}
		Dataset & Samples & Features & Accuracy: $k$-NN & Accuracy: $k$-PCN  \\
		\hline \hline
		Iris                  & 150   & 4   &  95.3 & 94.7 \\   \hline
		Wine                  & 178   & 13  &  90.9 & 89.7 \\   \hline
		Breast Cancer         & 569   & 30  &  95.9 & 93.3 \\   \hline
		Ionosphere            & 351   & 33  &  84.3 & 92.3 \\   \hline	
		Credit approval       & 689   & 35  &  75.3 & 79.3 \\   \hline				
	\end{tabular}
\end{table}

\section{Discussion}

While sparse regularization methods (e.g., involving a $\ell_1$ penalty) tend to dominate research in partial correlation networks, we have shown that dense estimates resulting from $\ell_2$-penalties or related methods have a number of interesting advantages and relationships.
For one, the dense network can be computed in a single direct matrix computation, rather than an iterative method or separate regression problem for every node. 
We also demonstrated how this could be performed with relative efficiency even for large datasets if the data was of much lower rank (or regularized to have a lower rank).
This approach could be subsequently used to produce a sparse network, for example by thresholding of edge weights. 
Or, as we have shown, the dense network itself has deep relationship to popular spectral embedding methods, suggesting a new interpretation and new avenue of extensions for such methods.

%
%
%
%
%

%
\bibliography{efficient_algorithm_01.bbl}
\bibliographystyle{plain}

\end{document}